\documentclass[11pt,a4paper]{article}
\usepackage{acl2019}
\usepackage{times}
\usepackage{latexsym}
\usepackage{url}

\aclfinalcopy
\title{Mapping supervised bilingual word embeddings from English to low-resource languages}

\author{Sourav Dutta \\
	Computational Linguistics and Phonetics \\
	Saarland University, Germany \\
	\texttt{souravd@coli.uni-saarland.de}}

\date{}

\begin{document}
\maketitle
\begin{abstract}
	It is very challenging to work with low-resource languages due to the inadequate availability of data. Using a dictionary to map independently trained word embeddings into a shared vector space has proved to be very useful in learning bilingual embeddings in the past. Here we have tried to map individual embeddings of words in English and their corresponding translated words in low-resource languages like Estonian, Slovenian, Slovakian, and Hungarian. We have used a supervised learning approach. We report accuracy scores through various retrieval strategies which show that it is possible to approach challenging tasks in Natural Language Processing like machine translation for such languages, provided that we have at least some amount of proper bilingual data. We also conclude that we can follow an unsupervised learning path on monolingual text data as that is more suitable for low-resource languages.
\end{abstract}

\section{Introduction}
Proper data is very crucial for any task related to natural language. Solving problems for data-rich languages like English, German, French, Italian, and Spanish is relatively much easier when compared to that of low-resource languages for which data is really scarce. Word embeddings play a very important role in various tasks of understanding, processing and generation of natural language, including but not limited to machine translation, text categorization, semantic understanding, and other relevant tasks. Research related to word embeddings started back in 2013 with Word2Vec \cite{mikolov2013distributed}. Significant breakthroughs were made in the subsequent years like GloVe \cite{pennington2014glove}, and FastText \cite{bojanowski2017enriching}. Recently there has been a lot of attention on bilingual word embeddings. Most of the methods to learn such vectors use some kind of bilingual values at the document level. These signals can be in form of aligned and comparable corpora. More commonly, they can also be in form of parallel corpora.

\noindent \citet{mikolov2013exploiting} talks about an alternative approach which involves first training word embeddings from monolingual corpus of each language and then mapping both of them to a common shared vector space with respect to a bilingual dictionary. The advantage of this approach is that it requires very minimal supervision compared to the other approaches mentioned before. Moreover, once we learn the mappings according to the bilingual dictionary, we can apply it to words which are unseen before or \emph{out-of-vocabulary (OOV)} words. This is actually very helpful for tasks which involve such unseen words during training, for example, machine translation.

\noindent \citet{artetxe2018aaai} follow this path and propose a multi-step framework that generalizes the previous work done in the field to map bilingual embeddings efficiently from monolingual corpora using orthogonal transformations of both word vectors. In our experiment, we apply this algorithm to map the trained word embeddings from English to four lower-resource European languages from their respective corpus to generate bilingual word embeddings for each of these languages from English as a source. Getting proper data for low-resource languages is a very challenging task, so here we use easily-available parallel corpora of these languages with English.

\noindent The remaining part of this paper is organized as follows. In section 2, we first talk about some other research work related to topic of mapping bilingual embeddings. Section 3 discusses the method which we have followed in this experiment. Section 4 then talks about the different steps of our experiment. In section 5, we then report our results and analyze how the approach works for low-resource languages. Section 6 and 7 conclude the paper and propose future scope of extending this work.

\section{Related Work}
Here, we will discuss some of the related work relevant to this topic. We know that mapping bilingual word embedding works by independently training the monolingual embeddings of two different languages. Then we map them into a shared vector space based on the word pairs in a bilingual dictionary. Let us briefly discuss some methods which try to achieve this objective.

\noindent \textbf{Canonical Correlation Analysis:} The objective here is to map the word embeddings into a shared space with the goal of maximizing the similarity between them. This was first proposed by \citet{faruqui2014improving} through Canonical Correlation Analysis (CCA). There are newer version of this approach based on the idea of the initial CCA method.

\noindent \textbf{Maximum Margin:} In this approach, the objective is to maximize the margin or difference between the correct translation and all the other candidates. This was proposed by \citet{dinu2014improving} where they used intruder negative sampling in order to generate more training samples so that they could tackle the \emph{hubness} \footnote{Hubness problem is faced when one tries to map higher dimensional data to a lower dimension, resulting in some words having multiple nearest neighbors in the target.} problem.

\noindent \textbf{Regression methods:} First proposed by \citet{mikolov2013exploiting}, the embeddings of one language are mapped to maximize their similarity scores with embeddings from another language. It follows the linear regression methodology with a least squares objective function that minimizes the Euclidean distance between the word pair entries in the bilingual dictionary. This method was later on adopted and improved by many other researchers, where it was mentioned that instead of learning the mapping from source to target words, doing it in reverse direction is a better way to tackle the hubness problem.

\noindent As we can see, there has already been a variety of work done related to mapping word embeddings with different ideas and approaches. However, the main objective behind working with multilingual or bilingual embedding mapping is to extract lexicons in both languages. This is mainly done by \emph{nearest neighbor retrieval} where the target word with the closest similarity score (mostly cosine) to the source word is picked. \citet{dinu2014improving} suggested to use \emph{inverted nearest neighbor retrieval} where the same algorithm is applied in the opposite direction in order to avoid the hubness problem in dimensional mapping of embeddings. Recently, \citet{smith2017offline} suggested \emph{inverted softmax retrieval} method where they reversed the embedding direction like before but instead of using cosine similarity they used a softmax function with hyperparameter tuning in the training dictionary. Table 1 shows the accuracy scores for bilingual mapping from English to the high-resource languages with each of these retrieval algorithms as reported in \citet{artetxe2018aaai}.

\begin{table*}[t]
	\begin{center}
		\begin{tabular}{|l|c|c|c|}
			\hline \textbf{Language} & \textbf{nn (\%)} & \textbf{inn (\%)} & \textbf{inv. soft. (\%)}\\ \hline \hline
			\textbf{EN-DE} & \textbf{44.27} & 42.20 & 44.13 \\
			\textbf{EN-IT} & 44.00 & 43.07 & \textbf{45.27} \\
			\textbf{EN-ES} & 36.53 & 32.53 & \textbf{36.60} \\
			\textbf{EN-FI} & \textbf{32.94} & 31.18 & \textbf{32.94} \\ \hline
		\end{tabular}
	\end{center}
	\caption{\label{font-table} Percentage accuracy scores of mapping bilingual embeddings for high-resource languages as mentioned by \citet{artetxe2018aaai} measured using different retrieval methods (\emph{nn}=Nearest Neighbor, \emph{inn}=Inverted Nearest Neighbor, \emph{inv. soft.}=Inverted Softmax).}
\end{table*}

\section{Method}
We first train word embedding models for both English and each of the low-resource languages we are working with, which is discussed more in next section. In our experiment, we follow the same steps as that of \citet{artetxe2018aaai}. We basically reuse their open-source software VecMap\footnote{VecMap: https://github.com/artetxem/vecmap} to train the mapping of bilingual word embeddings using the training word pair dictionary. We also use the same software to test the trained embeddings on the test data.

\noindent Here we will discuss in brief the basic idea behind the approach suggested by \citet{artetxe2018aaai}. There are multiple steps mentioned in their paper, but we will just mention the idea of \emph{orthogonal mapping} because it is responsible for learning from monolingual data and mapping the bilingual word embeddings to a common space. The other steps like \emph{whitening, re-weighting, de-whitening,} and \emph{dimensionality reduction} are optional. Details about those methods are available in their work. We have here worked with the most basic parameters involving just orthogonal mapping without any kind of prior preprocessing mentioned in their work. We have applied our own preprocessing methods to the raw data and then proceeded to train the bilingual word embeddings.

\noindent \textbf{Orthogonal Mapping:} This is the step that maps the monolingual word embeddings from the two source and target languages to a shared vector space. The dot product of each language is preserved using orthogonal SVD transformations, resulting in the maximal cross-variance of the mapped embeddings. This is how they train bilingual word embeddings from monolingual data. However, it also needs a training dictionary to refer to the word pairs in both languages and learn mappings of those words.

\noindent \textbf{Cross-domain similarity local scaling (CSLS):} Our goal is to produce matching word pairs between two languages. This means we want to use a comparison metric such that the nearest neighbor of a source word in the target language, is more likely to have this source word as its nearest neighbor. If \emph{y} is a K-NN (K-Nearest Neighbor) candidate of \emph{x}, it does not necessarily mean that it also holds true the other way around. \citet{conneau2017word} introduced a new comparison metric for this purpose which was termed \emph{Cross-domain similarity local scaling (CSLS)}. It is a measure of updating the cosine similarity score between two embeddings by deducting from it the mean similarity scores of a source embedding to its target and vice-versa. More details and mathematical representations about CSLS are mentioned in their work.

\noindent We report the normal nearest neighbor (nn) score as well as the CSLS metric accuracy. The main reason why we chose CSLS as a comparison metric in this case is CSLS increases the similarity associated with isolated word vectors. It shows significant increase in the overall efficiency of the word translation retrieval. On top of that, it does not require any additional parameter tuning as well.

\section{Experiment}
\citet{artetxe2018aaai} have applied their algorithm on German (DE), Italian (IT), Spanish (ES), Finnish (FI), all of which are rich in textual resources on the web. We have tried to apply the same method on low-resource languages like Estonian (ET), Slovenian (SL), Slovakian (SK), and Hungarian (HU). We have also run this experiment on the resource-rich languages mentioned above using our own preprocessing steps. The steps of our experiment are mentioned as follows, with detailed explanation of each step following hereafter in the next section.

\verb|-> Preprocessing|

\verb|-> Prepare word embeddings|

\verb|-> Prepare bilingual dictionary|

\verb|-> Train-test split|

\verb|-> Train bilingual embeddings|

\verb|-> Test accuracy|

\noindent Files generated in each step of the experiment are saved to the local disk through Python's \emph{pickle} library such that they can be used multiple times without spending the time to create them repeatedly. This enables us to dynamically save and load the operational files during runtime, making the overall experiment much faster and more efficient. The code for this experiment is open-source\footnote{https://github.com/SouravDutta91/map-low-resource-embeddings}.

\subsection{Data}
Getting proper text data for any low-resource language is a challenging task. As we want to address bilingual embeddings, we wanted to use parallel corpora because it ensures high probability of a word and its corresponding translated versions in all the languages being present in the corpora. This is the reason why we have used \textbf{Europarl} \cite{koehn2005europarl}, which is a collection of parallel text corpora from the proceedings of the European Parliament. It comprises of text versions in 21 different languages including Romanic, Germanic, Finni-Ugric, Baltic, Slavic, and Greek. Among these options, we have chosen to use the Europarl version of \emph{Estonian (ET)}, \emph{Slovenian (SL)}, \emph{Slovakian (SK)}, and \emph{Hungarian (HU)} as they can all be considered to be low-resource languages.

\noindent We prepare dictionaries of parallel words in English and the low-resource language (ET, SL, SK, HU). We then split the word pairs into train and test data, which we further use in the mapping algorithm.

\begin{table*}[t!]
	\begin{center}
		\begin{tabular}{|l|cccc|}
			\hline \textbf{Language} & \textbf{Total words} & \textbf{Preprocessed} & \textbf{Train} & \textbf{Test} \\ \hline
			\textbf{EN-ET} & 6249 & 6096 & 4267 & 1829 \\
			\textbf{EN-SL} & 8600 & 8329 & 5830 & 2499 \\
			\textbf{EN-SK} & 4111 & 3984 & 2789 & 1195 \\
			\textbf{EN-HU} & 8872 & 8460 & 5922 & 2538 \\
			\hline
		\end{tabular}
	\end{center}
	\caption{\label{font-table} Number of words in each language corpus; \emph{Preprocessed} refers to word frequency after preprocessing step; \emph{Train} and \emph{Test} refer to the number of words used in train and test steps respectively.}
\end{table*}

\noindent Table 2 shows a quantitative summary of the total number of translated word-pairs in each language pair, both before and after the preprocessing step. It also shows us the number of such word pairs used for training and testing phases. We discuss all the detailed steps of data processing in the next sections.

\subsection{Preprocessing}
The Europarl corpus contains European parliament proceedings data as a collection of sentences. As a result, it includes a variety of numbers, punctuation symbols and other non-alphabetic character representations. We first tokenize each sentence into its constituent words. One important thing to note here is that we ignore most of the punctuation symbols through regular expressions during this step so that we work with only words present in the corpora.

\subsection{Word Embeddings}
Here we use Word2Vec \cite{mikolov2013distributed} to train the word embeddings for our data. Once we have finished preprocessing, we convert the resulting words into a list of sentences and feed these files for each language into the Gensim toolkit \cite{vrehuuvrek2011gensim} to create the corresponding embedding model. For each target low-resource language, we have Word2Vec models for the source (English) and the target (that language).

\noindent For the training step, we consider only those words which have a minimum count of at least 5 in the entire corpus. We train the 300 dimensional embeddings with a window size of 5 over 10 iterations across multiple available processors. To increase efficiency of the process, we also include a negative skip-gram sampling of 5.

\subsection{Translation}
In order to train the mapping of embeddings from one language to another through a supervised approach, we need pairs of words in each both languages. These word pairs are needed not only for the training phase but also to test the accuracy of our experiment. However, as mentioned before, availability of proper data for low-resource languages is very rare. It is even more challenging to get a dictionary of word pairs in English and such languages. We have used the \textbf{Yandex Translation API} \footnote{Powered by Yandex.Translate - https://translate.yandex.com/} for this purpose. Although it is free to use, it should be noted that this service has a daily request limit of 1,000,000 characters and a monthly limit is 10,000,000 characters.

\noindent We feed the preprocessed English words into this service and obtain their corresponding translations in the respective low-resource language. Often the translation of certain words may result into multiple words or phrases in the target language. As we are dealing with word embeddings only, so we ignore such words whose translation results in multiple words or a phrase. All the word pairs obtained in this step are saved in a dictionary.

\subsection{Training and testing}
We generate training and testing data from the dictionary of word pairs that we obtained in the previous step. We split the total number of pairs in the saved dictionary into roughly 70\% for training and the remaining 30\% for testing. Both the train and test datasets are shuffled for randomness.

\section{Results and Discussion}
We have run this experiment on both low-resource as well as high-resource languages for comparison between the two groups. We report findings based on output parameters like coverage and accuracy scores. Accuracy can be reported with respect to different retrieval methods of comparison between the source and target bilingual word embeddings. In this experiment, we report accuracy scores based on the nearest neighbor (nn) and the CSLS retrieval modes.

\noindent \textbf{Coverage:} Coverage refers to the percentage of data covered or mapped by the algorithm during testing phase. Table 3 shows the coverage percentages for words in each of the target languages from English. Just like \citet{artetxe2018aaai}, we have 100\% coverage for all the four different high-resource languages. One the other hand, when we ran the same experiment on the low-resource languages, we get much lower coverage values between \textbf{25\% - 29\%}. The main reason behind this is the preprocessing step. We have performed a very basic step data preprocessing by just removing the digits, punctuation symbols and other similar characters. On top of this, they have further additional steps of preprocessing like \emph{whitening, re-weighting, normalization, de-whitening,} and \emph{dimensionality reduction}. This helps in creating a much better bilingual dictionary of proper words, which are used for training and testing. If we also follow these preprocessing steps, we will get much higher percentage scores of coverage for the low-resource languages.

\noindent \textbf{Accuracy:} Accuracy refers to the percentage of test data correctly mapped from the bilingual word embeddings. Table 4 shows a list of the accuracy scores for both categories of languages. Comparing the values from Table 1, we can see that the nearest neighbor accuracy scores for the high-resource languages (DE, IT, ES, FI) are very nearby. This is because we used the same bilingual dictionary and embedding files as that of \citet{artetxe2018aaai}. We also see that a significant increase (up to \textbf{5.33\%}) in the accuracy scores when we use CSLS retrieval mode. Coming to the low-resource languages (ET, SL, SK, HU), we see much lower accuracy scores compared to the other group of languages with nearest neighbor accuracy ranging between \textbf{11.07\% - 16.69\%}. The CSLS retrieval scores for these languages are again relatively higher than the nearest neighbor scores ranging between \textbf{12.42\% - 18.06\%}. It is expected that accuracy scores for the low-resource languages will be much less than those of the high-resource counterparts due to the higher amount of data available for the resource-rich languages. This is because with lesser amount of data (words), there is obviously lesser probability of proper words being present in the target language corpus for a particular source word.

\noindent One more important thing to note is that we have used the \emph{Yandex Translation API} to generate the dictionary of bilingual words, which is after all a neural machine translation system and so, it will not have a translation accuracy of 100\%. Although it is time-consuming and expensive, we can probably increase the accuracy score more if we use manual translation annotation by a native speaker of the language or an expert translator (for example, through methods of crowd-sourcing). We also had a good number of \emph{OOV} words for both the categories of languages, which tells us that this approach is probably a good option to look for when we want to want to tackle tasks like machine translation.

\begin{table}[t!]
	\begin{center}
		\begin{tabular}{|l|c|}
			\hline \textbf{Language} & \textbf{Coverage (\%)} \\ \hline \hline
			\textbf{EN-DE} & 100 \\
			\textbf{EN-IT} & 100 \\
			\textbf{EN-ES} & 100 \\
			\textbf{EN-FI} & 100 \\ \hline
			\textbf{EN-ET} & 26.35 \\
			\textbf{EN-SL} & 25.77 \\
			\textbf{EN-SK} & 24.94 \\
			\textbf{EN-HU} & 28.80 \\
			\hline
		\end{tabular}
	\end{center}
	\caption{\label{font-table} Percentage of words covered (mapped) successfully for both high-resource languages (DE, IT, ES, FI) and for low-resource languages (ET, SL, SK, HU) in our experiment. \citet{artetxe2018aaai} reported 100\% coverage as well.}
\end{table}

\begin{table}[t!]
	\begin{center}
		\begin{tabular}{|l|c|c|}
			\hline \textbf{Language} & \textbf{nn (\%)} & \textbf{CSLS (\%)} \\ \hline \hline
			\textbf{EN-DE} & 44.73 & \textbf{49.20} \\
			\textbf{EN-IT} & 43.07 & \textbf{48.40} \\
			\textbf{EN-ES} & 35.60 & \textbf{39.20} \\
			\textbf{EN-FI} & 32.02 & \textbf{36.10} \\ \hline
			\textbf{EN-ET}  & 11.83 & \textbf{16.18} \\
			\textbf{EN-SL} & 13.35 & \textbf{16.46} \\
			\textbf{EN-SK} & 11.07 & \textbf{12.42} \\
			\textbf{EN-HU} & 16.69 & \textbf{18.06} \\
			\hline
		\end{tabular}
	\end{center}
	\caption{\label{font-table} Percentage accuracy scores of mapping bilingual embeddings for low-resource and high-resource languages; \emph{nn} refers to Nearest Neighbor retrieval in our experiment; \emph{CSLS} refers to the accuracy scores when we use the CSLS algorithm for retrieval.}
\end{table}

\section{Conclusion}
Here we have tried to replicate the steps as described by \citet{artetxe2018aaai} to train individual monolingual word embeddings and then map them to a shared vector space to generate bilingual word embeddings between two languages, from English to a low-resource European language. We have seen that words from such languages can also be mapped to a shared vector space with English words following a supervised method. Compared to the 49.2\% CSLS retrieval accuracy  in case of English-German bilingual mapping, we achieve the highest score of \textbf{18.06\%} CSLS accuracy for English-Hungarian among all the low-resource languages. Considering these languages are resource hungry, and that we can probably improve our preprocessing methods, we can safely conclude that this approach works for low-resource languages just like their high-resource counterparts. We assume that unsupervised approaches of learning bilingual word embeddings for these family of languages should also give satisfactory results if we are able to feed in a good amount of proper monolingual corpus data of these languages. We can also probably conclude that the accuracy scores for both low- as well as high-resource languages should be somewhat in a closer range if the amount of training and testing data for each language is on a similar comparable scale.

\section{Future Work}
Here we have worked with a supervised approach trained on a dictionary of translated word pairs. Given that we are working with low-resource languages, it is more ideal to go for an unsupervised method because proper bilingual parallel data might not be readily available all the time for all languages. So, we can try to extend this work with an unsupervised approach similar to \citet{artetxe2018acl} over monolingual data (for example, extracted data from Wikipedia dumps \footnote{https://dumps.wikimedia.org/}).

\noindent We also saw that the amount of proper available data plays a very important role in the percentage of mapping accuracy for the bilingual word embeddings. So, we can try to collect more text data in the low-resource languages and rerun the same experiment, hoping to get relatively better results.

\noindent This work can be further extended to cover Polish, Latvian, Lithuanian, Romanian, and other similar low-resource languages, which are available in both Europarl versions as well as Wikidumps. Similarly, we can also look forward to work with similar rare languages in other regions like Latin America, Asia, or even of African origin applying unsupervised algorithms on monolingual data only.

\section*{Acknowledgments}

This research was conducted as a project report for the seminar \emph{Word Embeddings for NLP and IR}, Saarland University, Germany. We would like to sincerely thank \textbf{Dr. Cristina España i Bonet} for offering this seminar and guiding us whenever we needed help. \\

\bibliography{acl2019}
\bibliographystyle{acl_natbib}

\end{document}